\documentclass[conference]{IEEEtran}
\IEEEoverridecommandlockouts
\usepackage{cite}
\usepackage{amsmath,amssymb,amsfonts}
\usepackage{algorithmic}
\usepackage{graphicx}
\usepackage{textcomp}
\usepackage{xcolor}
\usepackage{makecell}
\usepackage{multirow}
\usepackage{threeparttable}
\def\BibTeX{{\rm B\kern-.05em{\sc i\kern-.025em b}\kern-.08em
    T\kern-.1667em\lower.7ex\hbox{E}\kern-.125emX}}
\begin{document}

\title{Active Learning for NLP with Large Language Models}

\author{\IEEEauthorblockN{Xuesong Wang}
\IEEEauthorblockA{\textit{Department of Electrical and Computer Engineering} \\
\textit{Wayne State University}\\
Detroit, MI, USA \\
xswang@wayne.edu}\thanks{This is the project report for ECE 7500 at Wayne State University for Fall 2023 semester.}
}

\maketitle

\begin{abstract}
Human annotation of training samples is expensive, laborious, and sometimes challenging, especially for Natural Language Processing (NLP) tasks. To reduce the labeling cost and enhance the sample efficiency, Active Learning (AL) technique can be used to label as few samples as possible to reach a reasonable or similar results. To reduce even more costs and with the significant advances of Large Language Models (LLMs), LLMs can be a good candidate to annotate samples. This work investigates the accuracy and cost of using LLMs (GPT-3.5 and GPT-4) to label samples on 3 different datasets. A consistency-based strategy is proposed to select samples that are potentially incorrectly labeled so that human annotations can be used for those samples in AL settings, and we call it mixed annotation strategy. Then we test performance of AL under two different settings: (1) using human annotations only; (2) using the proposed mixed annotation strategy. The accuracy of AL models under 3 AL query strategies are reported on 3 text classification datasets, i.e., AG's News, TREC-6, and Rotten Tomatoes. On AG's News and Rotten Tomatoes, the models trained with the mixed annotation strategy achieves similar or better results compared to that with human annotations. The method reveals great potentials of LLMs as annotators in terms of accuracy and cost efficiency in active learning settings.
\end{abstract}


\section{Introduction}

Supervised deep learning requires a large amount of ground truth labels. Usually the samples are labeled by humans. However, annotation by human annotators is expensive, laborious, and sometimes challenging, which is especially true in Natural Language Processing (NLP) systems \cite{R1,R2,R3}. On Google Cloud Platform\footnote{https://cloud.google.com/ai-platform/data-labeling/pricing}, the text classification labeling costs for 1000 units (50 words per unit) \$129 in Tier 1 and \$90 for Tier 2. In \cite{R3}, the authors assumes \$0.11 per 50 tokens, and I will follow the same estimates in this work.

To reduce the labeling cost and enhance the sample efficiency, Active Learning (AL) technique can be utilized, assuming not all samples are equal \cite{R1}. AL algorithms start the model training using a few initially labeled samples, iteratively select most representative and(or) informative samples to be labeled by annotator, add the newly-labeled samples into the training set to continue training or train from scratch, and stop until stopping criteria is met. It can achieve similar or even better results compared to passive training which trains the model using the entire training set \cite{R2}, but the labeling cost is much lower.

To reduce the annotation cost, some researchers have proposed to use Large Language Models (LLMs), e.g., GPT-3, to label the instances thanks to the few-shot (in-context) learning capability of LLMs \cite{R3}. It costs 50\% to 96\% less to use labels from GPT-3 than using labels from humans to reach the same performance on multiple NLU and NLG tasks \cite{R3}. It is found that data labeled by GPT-3 is usually more noisy than human-labeled data, and the authors in \cite{R3} use the confidence score of GPT-3 API to measure the uncertainty of the model output. If the output confidence score is low, the sample will be labeled by human instead. As a result, the samples that labeled by GPT-3 and humans are combined to train the model to achieve a better performance with the limited labeling budget. When using LLMs to label the instances, the demonstration in the prompt plays an important role in the accuracy of the annotation. Authors of \cite{R4} investigated effects of the number of examples in the demonstration and the strategies to select examples.

In order to explore the performance of using latest LLMs to label instances for AL in order to reduce even more costs of labeling, this paper utilizes GPT-3.5 and GPT-4 to label training samples. Two sets of experiments are designed. First, GPT-3.5 and GPT-4 are utilized to label samples in 3 datasets using different number of examples in the demonstration and different demonstration example selection strategies. Because the GPT-3.5 and GPT-4 APIs do not have the confidence score output parameter anymore, a consistency-based strategy is proposed to find the samples that the GPT-3.5 or GPT-4 is not confident in. The annotation accuracy, actual cost and the inconsistency rate are reported. The second set of experiments are using GPT-3.5 in AL experiments. For those samples that GPT-3.5 are uncertain, we use human annotations; the GPT-3.5 annotations are used for the rest of the samples. We report the accuracy of AL experiments and compare the model performance between the models using human annotations only and mixed annotations.

The paper is organized as follows: the related literature is briefly reviewed in section~\ref{related_work}, followed by the detailed introduction of the proposed method in section~\ref{method}. In section~\ref{experiment_setup}, the experiment setup is given, and the experiment results are illustrated and analyzed in section~\ref{results}. Finally, the limitation, future work are discussed and the conclusion is drawn in section~\ref{conclusion}.

\section{Related Work}\label{related_work}

A general procedure of AL is illustrated in Table~\ref{active_learning} \cite{R1}. Most of the literature focus on the query strategy design. One dimension is informativeness of the samples. For instance, we can select samples based output uncertainty, like Least Confidence \cite{R5}, Prediction Entropy \cite{R6}, and Breaking Ties \cite{R7}. We can also use multiple models to select the most disagreed samples \cite{R8}. Instances that would most strongly impact the model should be considered informative, therefore, we can use the norm of gradient to select the instances \cite{R9}. Considering informativeness only might introduce sampling bias and the selection of outliers, thus the representativeness should be taken into consideration as well \cite{R1}. For instance, we can select instances that are more representative of the unlabeled set \cite{R10,R11}, or instances that are different from labeled instances \cite{R12,R13,R14}. Moreover, we can take a hybrid approach to combine the two metrics together \cite{R15,R16}. Authors of \cite{R2} compared 5 query strategies on 5 text classification datasets using 2 transformer models. In this work, 3 of those 5 query strategies are compared, i.e., random (as baseline), Least Confidence \cite{R5}, and Breaking Ties \cite{R7}, on 3 of the 5 datasets, i.e., AG's News \cite{R17}, Rotten Tomatoes Movie Reviews \cite{R18}, and TREC-6 \cite{R18}, using 1 transformer model, i.e., DistilRoBERTa \cite{R19}.

Using GPT-3 to label instances was proposed in \cite{R3} with a price of \$0.04 per 1k tokens \cite{R3}. To reduce the labeling noise in the GPT-3 annotation, they proposed to measure the model uncertainty using API output parameter $logprobs$. If $logprobs$ is low, the sample will be labeled by humans instead of using GPT-3 label. The labels created by GPT-3 and humans are mixed to train the model. They found that it costs 50\% to 96\% less to use the labels from GPT-3 than using labels from humans on a variety of NLU and NLG tasks to reach the same performance. The goal of this work is to test the latest GPT-3.5 and GPT-4 models, whereas the $logprobs$ output parameter is not present anymore. A consistency-based approach is proposed and introduced in \ref{method}.

Benefit from few-shot (in-context) learning capability of GPT-3.5 and GPT-4 models, LLMs can be used to annotate samples directly without finetuning, with just some demonstration examples in the prompt \cite{R4}. Authors of \cite{R4} explored the effects of demonstration example selection strategies, and they found that selecting the examples that are most similar to the test sample can reach the highest accuracy. In this work, three strategies are compared on GPT-3.5 and GPT-4 models as well, i.e., $random$ (as baseline), examples that cost fewest tokens ($min\_token$), and examples that are most similar to the test examples ($max\_similarity$).

\begin{table}[tb]
    \centering
    \caption{General Procedure of Active Learning}
    \begin{tabular}{l}
        \hline
        \emph{Input:} An unlabeled data pool $U$\\
        \emph{Output:} Final labeled dataset $L$ and trained model $M$\\
        \emph{Procedure:}\\
        1: Initialize $L$ and $U$\\
        2: train $M$ using $L$\\
        3: while not stop\_criterion() do:\\
        4: \hspace{4pt} samples $I$ $\leftarrow$ query($M$,$U$,$L$)\\
        5: \hspace{4pt} annotate samples $I$\\
        6: \hspace{4pt} $U = U - I$, $L = L + I$\\
        7: \hspace{4pt} train $M$ using $L$\\
        \emph{return:} $L$, $M$\\        
        \hline
    \end{tabular}
    \label{active_learning}
\end{table}

\section{Method}\label{method}

In this section, the annotation method using GPT-3.5 and GPT-4 will be introduced. Then consistency-based uncertain sample selection strategy is proposed.

\subsection{Annotation by GPT-3.5 and GPT-4}

The OpenAI Chat Completion API\footnote{https://platform.openai.com/docs/api-reference/chat} is used for the annotation because the legacy completion API is retiring. The pricing of the GPT-3.5 and GPT-4 models is listed in Table~\ref{pricing}. It requires a list of messages in the request, and returns a message, as shown in Table~\ref{messages}. The system prompt should be tuned for each different dataset in terms of the type of the texts, number of categories, and the class names. The API has a parameter called $response\_format$ which can be used to ensure the JSON format of the response. As we can see, $N*C$ demonstration examples are included in the messages given by the role \textbf{user}, where $N$ is the number of examples of each class included in the demonstration, and $C$ is the number of classes. Given a pool of labeled examples, we have different strategies to select a subset of them to be included in the demonstration. We can randomly select $N$ samples for each class ($random$), select the $N$ samples that consume fewest tokens ($min\_token$), or select $N$ samples that are most similar to the sentence to label ($max\_similarity$).

\begin{table}[tb]
    \centering
    \caption{Pricing of GPT-3.5 and GPT-4}
    \begin{tabular}{ccc}
        \hline
        Model & Input & Output\\
        \hline
        gpt-3.5-turbo-1106 & \$0.0010 / 1K tokens & \$0.0020 / 1K tokens\\
        gpt-4-1106-preview & \$0.01 / 1K tokens & \$0.03 / 1K tokens\\
        \hline
    \end{tabular}
    \label{pricing}
\end{table}

\begin{table}[tb]
    \centering
    \caption{Example Annotation Process using OpenAI Chat Completion API}
    \begin{tabular}{l|l}
        \hline
        Role & Content\\
        \hline
        \multicolumn{2}{l}{Input Messages}\\
        \hline
        system & \makecell[l]{You have been trained to classify movie reviews.The user\\will give you some movie reviews, and you will classify \\the reviews into 2 categories: Negative, Positive. \\You should use the given class names only. The examples \\are given by the user and they should be used as your\\ references. You should respond in a json format with\\the class names only.}\\
        user & 1. first example sentence; 2. second example sentence;\\
        user & 1. Negative; 2. Positive;\\
        user & 1. sentence to label; 2. sentence to label;\\
        \hline
        \multicolumn{2}{l}{Output Messages}\\
        \hline
        assistant & 1. Negative; 2. Positive;\\
        \hline
    \end{tabular}
    \label{messages}
\end{table}

\subsection{Consistency-base Uncertain Sample Selection Strategy}

Unlike GPT-3 completion API which is retiring, the Chat Completion API does not have the output parameter $logprobs$ to indicate the confidence of the model output. I propose to use a consistency-base method to measure the model uncertainty on the output samples. Two parameters, i.e., $n$ and $temperature$ are tuned to let the API return $n$ results with variations controlled by $temperature$. Larger $temperature$ creates more variations, and larger $n$ consumes more tokens. If there is any inconsistency among the $n$ results on any samples, those samples will be marked as \emph{inconsistent} and when we calculate the accuracy of the annotation, these samples are not counted. Over the active learning process, the inconsistent samples will be labeled by humans.

\section{Experiment Setup}\label{experiment_setup}

The first experiment set is designed to test and analyze the accuracy and the cost of GPT-3.5 and GPT-4 to label instances. Different number of examples are included in the demonstration and different demonstration example selection strategies are compared. The second experiment set compares the effects of GPT-3.5 annotator on the model performance in active learning settings.

\subsection{Experiment Set 1: Annotation by GPT-3.5 and GPT-4}

The set of experiments are conducted on 3 text classification datasets: AG's News \cite{R17}, TREC-6 and Rotten Tomatoes Movie Reviews \cite{R18}. For each dataset, 125 samples are randomly selected in a stratified manner, i.e., keep the ratio of the samples across categories. And a subset of stratified 25 samples over the 125 samples are taken as the initial set, from which the demonstration examples will be selected. We report the accuracy and cost on the other 100 samples. Because all the datasets have ground truth labels for training, validation and test sets, we can use the ground truth to measure the accuracy. To test the effect of the number of examples in demonstration, we tested $N={1,2,3,4,5}$. In order to determine how I should select the demonstration examples from the initial set, three demonstration example selection strategies are compared: $random$, $min\_token$, and $max\_similarity$. When calculating the cost in USD(\$), we use the pricing in Table~\ref{pricing} and the consumed input and output token numbers in the API's output. Due to the budget limitation, all the experiment results are reported over 1 run, instead of multiple runs. To measure the inconsistency, we let API return $n=3$ results, and $temperature$ is set to 0.2. We also report the inconsistency rate which indicates in the active learning settings, how many samples need to be labeled by humans.

\subsection{Experiment Set 2: Active Learning using Mixed Labels}

The set of experiments are designed to compare the performance of active learning using human annotations only and using mixed annotations from GPT-3 and humans. The same 3 datasets are used in this experiment set. The model trained in active learning process is DistilRoBERTa \cite{R19}, and 3 query strategies are tested, i.e., $random$, $Least\ Confidence$ \cite{R5} and $Breaking\ Ties$ \cite{R7}. All the layers are finetuned from pretrained model and the model is re-trained after each query iteration. The mean accuracy on test set and the mean Area Under Curve (AUC) of the test accuracy curve are reported over 3 runs. The model training has an early stop policy which will stop the model training if the model does not improve performance over a few epochs. For each experiment, we query 10 times, and 50 samples each query, totally 500 samples. 10\% of the labeled samples are used as validation set during model training. The initial set contains 25 samples, which are randomly stratified sampled instances. For GPT-3.5 labeler, $N$=3 for each class, and the demonstration selection strategy is $min\_token$.

\subsection{Datasets}

The 3 datasets used in this paper are AG's News \cite{R17}, TREC-6 and Rotten Tomatoes Movie Reviews \cite{R18}. The AG's News dataset is to classify the news topics into 4 categories, i.e., World, Sports, Business, and Sci/Tech. The dataset contains 120k samples in the training set, and 7600 in test set. The TREC-6 dataset is to classify questions into 6 coarse labels, i.e., Abbreviation, Entity, Description, Human, Location, and Numeric. The training set contains 5452 samples and the test set contains 500 samples. The Rotten Tomatoes dataset is to classify movie reviews into negative and positive classes. The training set contains 9596 samples and the test set contains 1066 samples. All the pre-processing procedure is aligned with \cite{R2}.

\section{Results}\label{results}

In this section, the results of two sets of experiments are discussed and analyzed.

\subsection{Results of Experiment Set 1}

\subsubsection{Cost Efficiency}

\begin{figure*}
    \centering
    \includegraphics[width=0.8\linewidth]{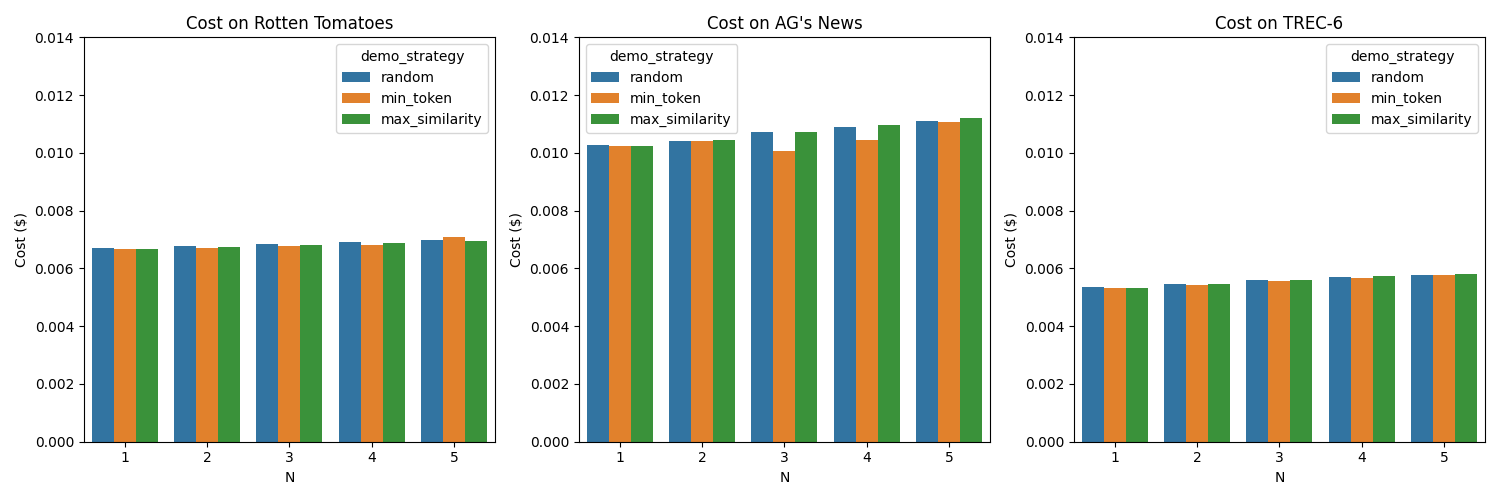}
    \caption{Cost of labeling 100 samples by GPT-3.5}
    \label{cost_gpt_3_5}
\end{figure*}

\begin{figure*}
    \centering
    \includegraphics[width=0.8\linewidth]{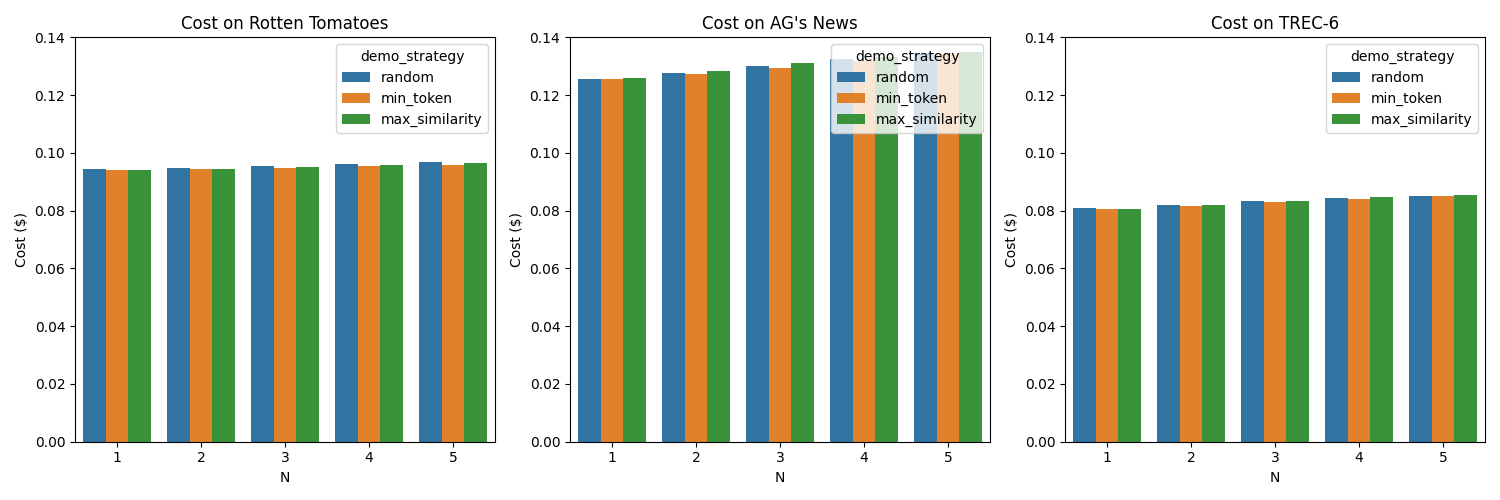}
    \caption{Cost of labeling 100 samples by GPT-4}
    \label{cost_gpt_4}
\end{figure*}

The costs of labeling 100 samples using GPT-3.5 and GPT-4 are shown in Fig~\ref{cost_gpt_3_5} and Fig~\ref{cost_gpt_4}, respectively. And the costs of labeling such 100 samples by human annotators are listed in Fig~\ref{cost_human}, assuming the same pricing listed in \cite{R3}. We can see that the labeling cost by GPT-3 is about 1/1000 of human annotations. And the cost of GPT-4 is about 10 times of GPT-3.5. Providing more examples in the demonstration consumes more tokens, thus causing more costs. In addition, we can see that $min\_token$ demonstration selection strategy costs less than or equal to other strategies.

\begin{table}[tb]
    \centering
    \caption{Cost of Labeling 100 Samples by Humans}
    \begin{tabular}{cccc}
        \hline
        Dataset & Rotten Tomatoes & AG's News & TREC-6\\
        \hline
        Cost (\$) & 11.02 & 12.16 & 11.00\\
        \hline
    \end{tabular}
    \label{cost_human}
\end{table}

\subsubsection{Accuracy}

\begin{figure*}
    \centering
    \includegraphics[width=0.8\linewidth]{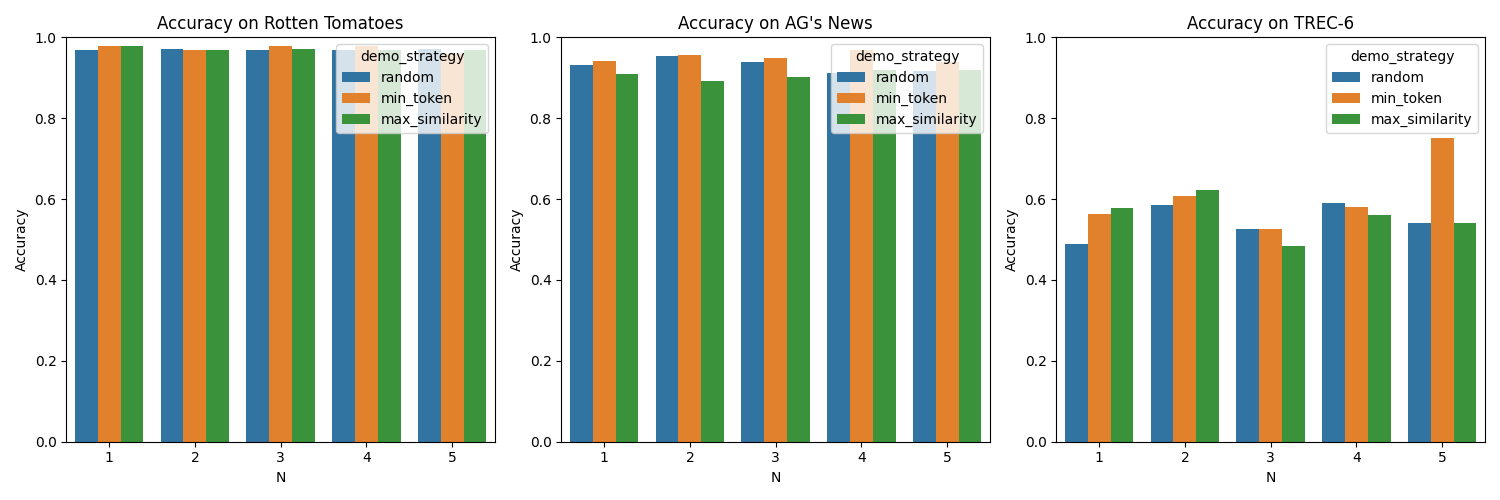}
    \caption{Accuracy of labeling 100 samples by GPT-3.5}
    \label{acc_gpt_3_5}
\end{figure*}

\begin{figure*}
    \centering
    \includegraphics[width=0.8\linewidth]{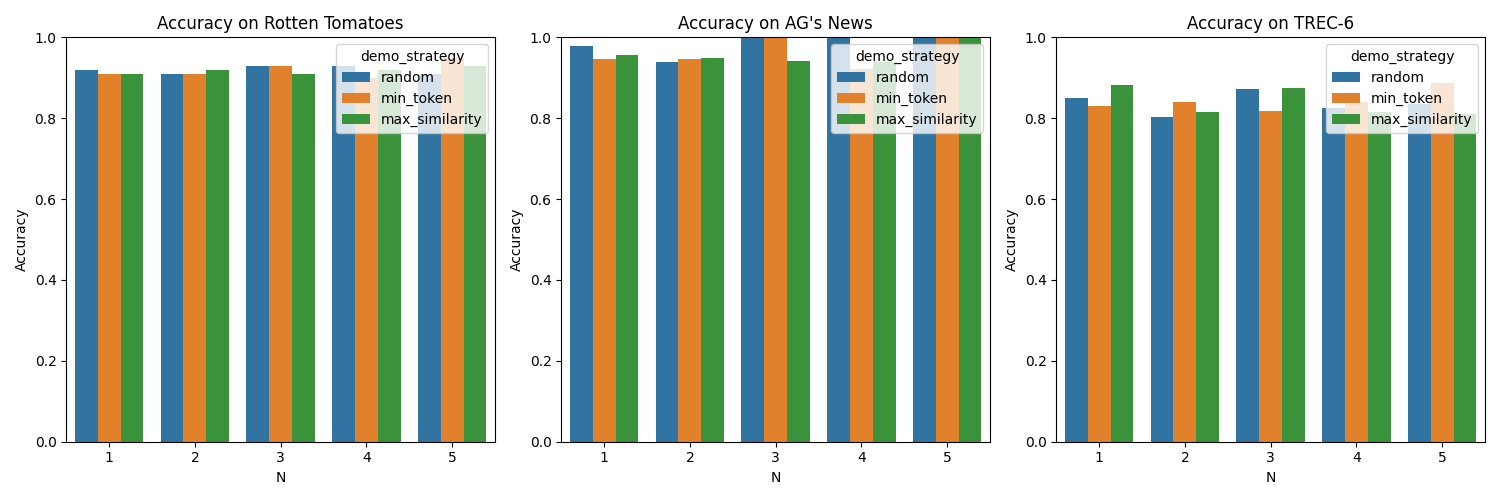}
    \caption{Accuracy of labeling 100 samples by GPT-4}
    \label{acc_gpt_4}
\end{figure*}

The annotation accuracy of GPT-3.5 and GPT-4 on 100 examples are shown in Fig~\ref{acc_gpt_3_5} and Fig~\ref{acc_gpt_4}, respectively. We can see that on AG's News and TREC-6, GPT-4 has higher accuracy, sometimes even close to 100\% accurate; however, on Rotten Tomatoes, GPT-4 has slightly lower accuracy compared to GPT-3.5, which deserves further investigation. The improvement on TREC-6 is significant, which indicates that on some difficult dataset, GPT-4 might have greater potentials. In terms of the number of demonstrations for each class $N$, I do not see a definite trend in the results. Among 3 demonstration selection strategies, $min\_token$ achieves highest or close to highest accuracy.

\subsubsection{Inconsistency Rate}

\begin{figure*}
    \centering
    \includegraphics[width=0.8\linewidth]{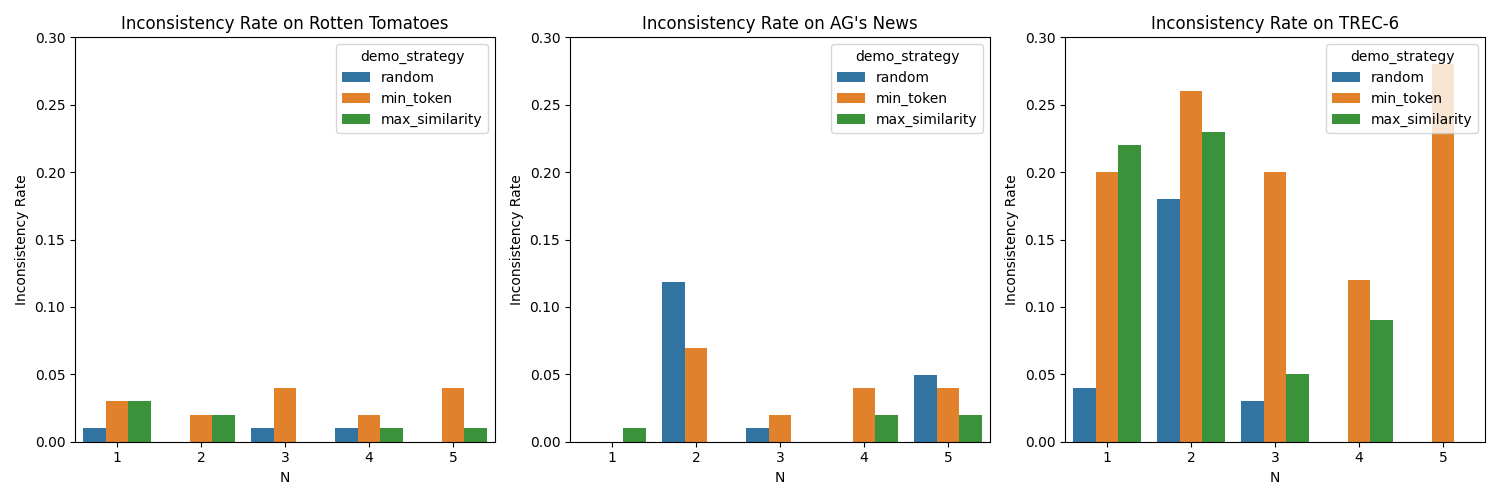}
    \caption{Inconsistency rate of labeling 100 samples by GPT-3.5}
    \label{inconsist_rate_3_5}
\end{figure*}

\begin{figure*}
    \centering
    \includegraphics[width=0.8\linewidth]{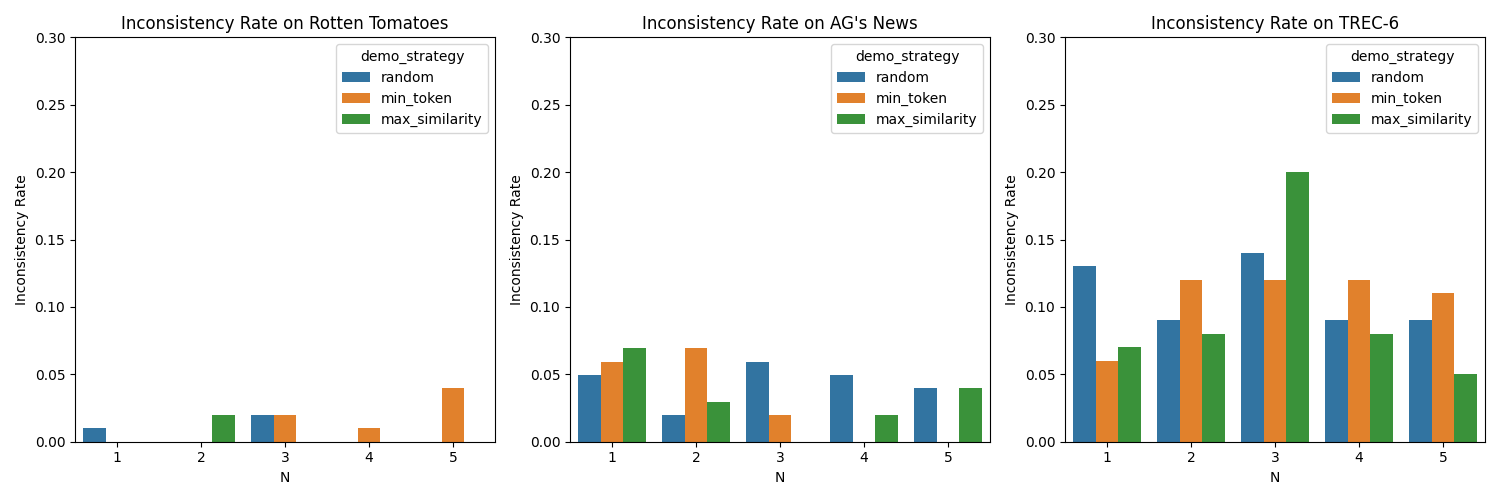}
    \caption{Inconsistency rate of labeling 100 samples by GPT-4}
    \label{inconsist_rate_4}
\end{figure*}

The inconsistency rates using GPT-3.5 and GPT-4 are shown in Fig~\ref{inconsist_rate_3_5} and Fig~\ref{inconsist_rate_4}, respectively. We can see that on TREC-6, GPT-3.5 has much higher inconsistency rate than GPT-4, indicating that it requires more human annotation if GPT-3.5 is used in the mixed labeling strategy. On the other datasets, GPT-4 has slightly lower inconsistency rate than GPT-3.5. We should note that the model showing inconsistency is better than making $consistent$ mistakes where multiple responses make the same mistakes. For $consistent$ mistakes, we cannot detect them since we do not have the ground truth for them during active learning settings. I do not see a correlation between the number of demonstration examples $N$ and the inconsistency rate; however, the $min\_token$ strategy has highest or close to highest inconsistency rate, which coincides with the findings in \cite{R4}.

\subsection{Results of Experiment Set 2}

The test accuracy and AUC of the test accuracy curve in active learning experiment sets are shown in Table~\ref{test_acc_auc}. As we can see that for AG's News and Rotten Tomatoes, the mixed labels from GPT-3.5 and humans can reach similar or even better results compared to using human annotations only. However, on TREC-6, the accuracy is significant lower than human annotation trained models, which might be explained by the low accuracy of the GPT-3.5 annotations. Among 3 query strategies, Breaking Ties achieved the best accuracy, but very close to Least Confidence, which is aligned with \cite{R2}.

\begin{table}[tb]
    \centering
    \caption{Test Accuracy and AUC in Active Learning}
    \begin{threeparttable}
    \begin{tabular}{*{6}c}
        \hline
        \multirow{2}{*}{Dataset} & Query & \multicolumn{2}{c}{Test Accuracy} & \multicolumn{2}{c}{Test Acc AUC}\\
        & Strategy & Human & Mixed & Human & Mixed\\
        \hline
        \multirow{3}{*}{AG} & BT & 0.90 & 0.88 & 0.85 & 0.84\\
                                   & LC & 0.89 & 0.87 & 0.82 & 0.82\\
                                   & R & 0.88 & 0.87 & 0.84 & 0.83\\
        \hline
        \multirow{3}{*}{RT} & BT & 0.84 & 0.81 & 0.76 & 0.77\\
                                         & LC & 0.83 & 0.84 & 0.76 & 0.78\\
                                         & R & 0.82 & 0.83 & 0.78 & 0.78\\
        \hline
        \multirow{3}{*}{TREC-6} & BT & 0.94 & 0.74 & 0.85 & 0.69\\
                                & LC & 0.94 & 0.70 & 0.81 & 0.65\\
                                & R & 0.82 & 0.75 & 0.82 & 0.67\\
        \hline
    \end{tabular}
    \begin{tablenotes}
        \item[1] AG: AG's News, RT: Rotten Tomatoes Movie Review
        \item[2] BT: Breaking Ties, LC: Least Confidence, R: Random
    \end{tablenotes}
    \end{threeparttable}
    \label{test_acc_auc}
    
\end{table}

\section{Discussion and Conclusion}\label{conclusion}

In this paper, we have explored the accuracy and cost efficiency of GPT-3.5 and GPT-4 as sample annotators for active learning. Through the first set of experiments, we have found that GPT-3.5 has lower cost and reasonable accuracy on easy tasks; on the contrary, GPT-4 has higher cost and performs better on harder tasks, e.g., TREC-6 question classification. With the proposed consistency-based label fixing strategy, we have demonstrated the potential of the mixed annotations over the human annotations. The experiments have shown that by using GPT-3.5 annotation and fixing incorrectly labeled samples with human annotation, the accuracy of the active learning models can be close to or even higher than human label trained models on easy tasks, e.g., movie review and news topic classifications. This work has offered insights for future dataset annotation process that GPT-3.5 or GPT-4 can be the initial annotator to reduce the cost, without losing too much accuracy.

Despite the interesting results the experiments have given, caution should be taken when using the proposed method. We should note that some datasets might be included in the training set of GPT-3.5 and GPT-4, therefore, they might not perform as well as shown in this paper when inference on new datasets. The reason why GPT-3.5 and GPT-4 behaves worse on TREC-6 deserves further investigation. The consistently mislabeled samples are not detectable in this work. A possible solution might be to randomly select a few samples from those samples that we think GPT-3.5 or GPT-4 label correctly, and label them by human efforts. However, this might cause higher costs. But anyways, this is all about trade-off between costs and accuracy.

\bibliographystyle{IEEEtran}
\bibliography{refs}
\end{document}